\documentclass[lettersize,journal]{IEEEtran}
\usepackage{amsmath,amsfonts}
\usepackage{algorithmic}
\usepackage{algorithm}
\usepackage{array}
\usepackage[caption=false,font=normalsize,labelfont=sf,textfont=sf]{subfig}
\usepackage{textcomp}
\usepackage{stfloats}
\usepackage{url}
\usepackage{verbatim}
\usepackage{graphicx}
\usepackage{cite}
\usepackage{mathrsfs}
\usepackage{multirow}
\usepackage{makecell}
\usepackage{booktabs}
\usepackage{hyperref}

\begin{document}

\title{Robust Multi-Model Fitting through Learning Neighbor Regions}

\author{Chang Nie, Guangming Wang, Zhe Liu,~\IEEEmembership{Member,~IEEE,} and Hesheng Wang,~\IEEEmembership{Senior Member,~IEEE,}
\thanks{This work was supported in part by the Natural Science Foundation of China under Grant 62225309, U24A20278, 62361166632 and U21A20480. (Corresponding Author: Hesheng Wang, e-mail: wanghesheng@sjtu.edu.cn)}%
\thanks{Chang Nie, Zhe Liu and Hesheng Wang are with School of Automation and Intelligent Sensing, Shanghai Jiao Tong University and Key Laboratory of System Control and Information Processing, Ministry of Education of China, Shanghai 200240, China.}%
\thanks{Guangming Wang is with the Department of Engineering, Cambridge
University, Cambridge CB2 1TN, UK.}}

\markboth{Journal of \LaTeX\ Class Files,~Vol.~14, No.~8, August~2021}%
{Shell \MakeLowercase{\textit{et al.}}: A Sample Article Using IEEEtran.cls for IEEE Journals}

\maketitle

\begin{abstract}
  Multi-model fitting involves fitting multiple models accurately in a noisy environment. It is the basis for computer vision tasks such as scene reconstruction and mixed reality. However, its performance is often limited by insufficient feature utilization, inefficient optimization, model overlap, and the non-differentiable pipelines. To overcome these limitations, we introduce a robust coarse-to-fine framework called Learning Neighbor Regions (LNR). Recognizing that substantial computational resources are wasted on numerous bad minimum sets, we propose the coarse-level module. This module utilizes a neural network to extract and analyze geometric feature of both local point-wise relationships and global contextual information in minimum sets, outputting confidence to pre-select a small number of good minimum sets, thereby enhancing overall efficiency before solving hypotheses. To address model overlap, LNR encodes neighbor region features for each hypothesis in its fine-level module. These region features consist of geometric features of neighboring data points, which can be used by multiple regions simultaneously. This design allows the neural network to individually refine and score each hypothesis. Importantly, LNR is trained to learn directly from data point features rather than from the hypothesis parameters, thus avoiding differentiating the sampling process and the model solvers. Extensive experiments on four classic multi-model fitting tasks demonstrate that LNR achieves state-of-the-art performance. The analysis suggests that LNR can be easily adapted to various robust multi-model fitting tasks. The code will be available at \url{https://github.com/IRMVLab/LNR}.
\end{abstract}


\section{Introduction}
\label{sec:intro}

\IEEEPARstart{R}{obust} multi-model fitting is an essential and widely applied task in computer vision. It involves identifying and fitting multiple models within a dataset, often complicated by significant noise. Improvements in this area have broad implications, benefiting numerous applications such as robot navigation \cite{kong2024robodepth, jian2022putn, jian2023path, li2023textslam}, 3D scene reconstruction \cite{rempe2021humor, guo2022neural, liu2022planemvs, yin2022towards}, autonomous driving \cite{jin2024multi, yang2023fast, liu2021vapid, yang2016fast, chang2018deepvp}, and mixed reality \cite{shi2024effective, jiang2024robust, chen2024pgsr, ebner2022video}. To address this task, a straightforward idea is to extend the single-model methods, like RANdom SAmple Consensus (RANSAC) \cite{fischler1981random}, to multi-model. For example, Sequential RANSAC \cite{vincent2001detecting} finds multiple instances by iteratively removing inliers. This greedy strategy resembles clustering, where each data point is assigned to a single model. Consequently, it may overlook the model overlap, where data points belong to both models. This limitation can lead to an underestimation of certain models, particularly when instances are not clearly separated.

\begin{figure}[t]
  \centering
   \includegraphics[width=0.99\linewidth]{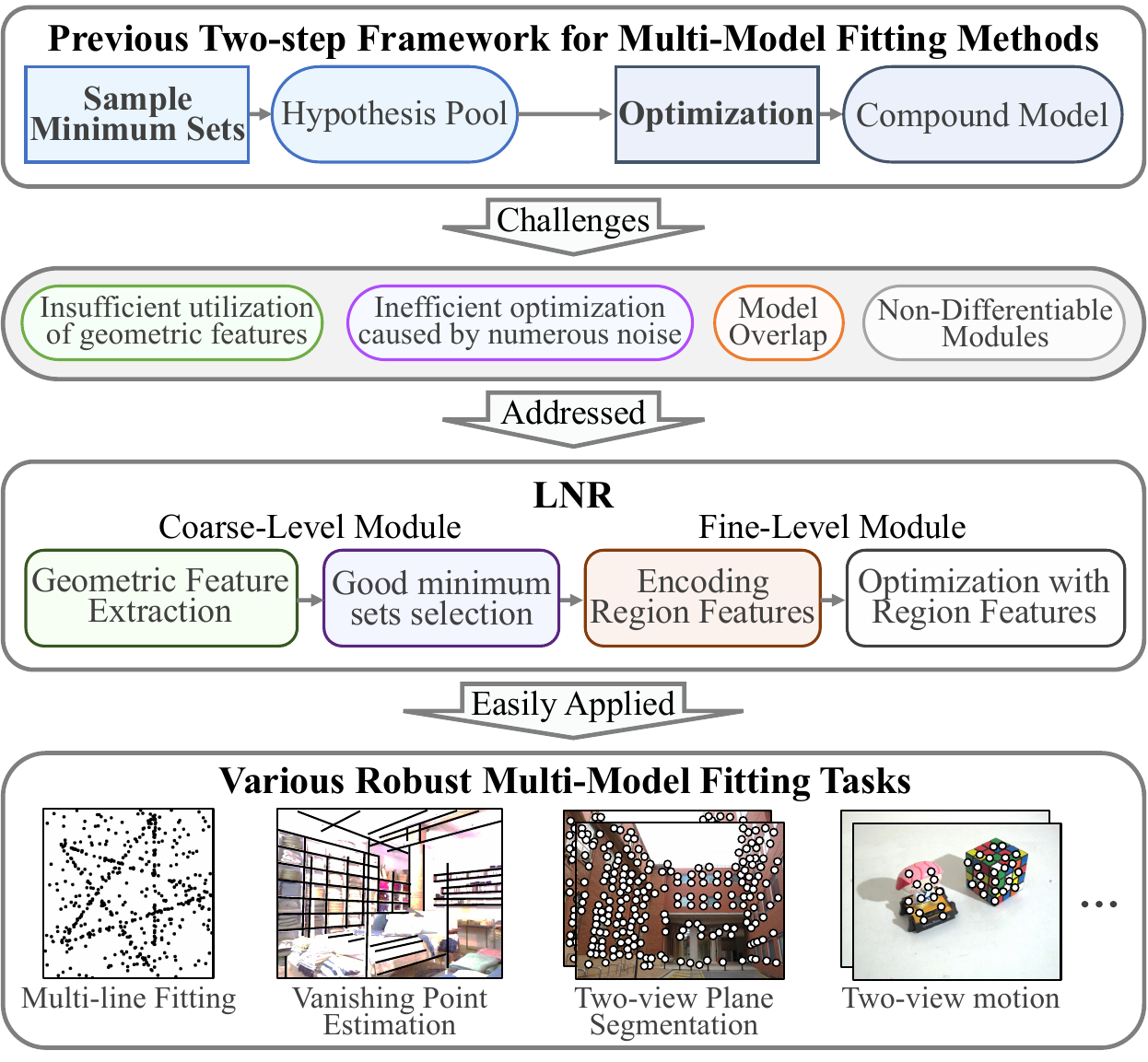}
    \vspace{-20pt}
   \caption{\textbf{LNR addresses the challenges of the two-step framework.} LNR uses neural network in the coarse module to fully extract geometric features and select good minimum sets for efficiency. In the fine-level module, LNR encodes region features for each hypothesis independently to deal with model overlap. This framework learns on data points rather than the hypothesis parameters, avoiding sampling process and model solvers}
   \vspace{-8pt}
   \label{fig:overall}
\end{figure}

Most recent methods adopt a two-step process. Firstly, they generate a hypothesis pool from minimum data sets \cite{barath2019progressive, barath2023finding, kluger2020consac, lu2021event, magri2021multilink}. These methods often employ iterative sampling techniques to expand the hypothesis pool, attempting to discard less promising ones. However, because each new model is built upon previous iterations, \textit{errors can accumulate due to model overlap} throughout the process. \textit{Numerous noise also causes inefficient optimization} on bad hypotheses. Secondly, a compound model is selected from the generated hypotheses \cite{isack2012energy, amayo2018geometric, barath2018multi, wang2015mode, farina2023quantum, feng2023hypergraph, levinkov2022higher}. Techniques include energy optimization \cite{isack2012energy, barath2018multi} or hyper-graph partitioning \cite{wang2015mode, feng2023hypergraph}. These methods often rely on complex cost functions to handle model overlap. These intricate cost functions, however, \textit{might not fully exploit the underlying geometric information} inherent in the data, potentially limiting their effectiveness across diverse and complex scenarios. These limitations motivate the exploration of alternative approaches that can more effectively utilize geometric features, such as integrating deep learning methods \cite{kluger2024parsac, kluger2020consac}. However, \textit{challenges arise in incorporating learning-based methods due to the non-differentiable} nature of traditional sampling and model-fitting processes.

To address the challenges of insufficient features utilization, inefficient optimization, model overlap, and the non-differentiable pipelines as shown in Fig. \ref{fig:overall}, we propose Learning Neighbor Regions (LNR) for robust multi-model fitting. To avoid error accumulation from iterative sampling, LNR simultaneously generates a large number of minimum sets through random sampling. This randomized approach ensures a broad coverage of potential models and enhances generalization. Meanwhile, to fully utilize the geometric features and manage the computational cost associated with numerous bad minimum sets, LNR introduces a coarse-level module. This module analyzes the geometric features of each minimum set to select good minimum sets before solving for the hypotheses.

For confusion caused by model overlap and the barrier of non-differentiable pipelines, LNR encodes neighbor region features for each hypothesis. This allows individual data points to contribute to multiple region features simultaneously, providing rich information to deal with various noises and model overlap. A neural network then analyzes these region features to refine and score the hypotheses. Importantly, LNR focuses on features derived from the data points, rather than the hypothesis parameters, avoiding the need to differentiate the sampling process or model solvers. Finally, to determine the number of models without prior knowledge, LNR employs non-maximum suppression (NMS), a post-processing technique commonly used in object detection, to output the final multi-model fitting results.

We evaluate LNR on four established multi-model fitting tasks: multi-line fitting, vanishing point estimation, two-view plane segmentation, and two-view motion estimation. Experimental results demonstrate that LNR achieves state-of-the-art performance across these tasks.

The main contributions of LNR are as follows:
\begin{itemize}
	\item We propose Learning Neighbor Regions (LNR), a novel framework for robust multi-model fitting. For challenges of non-learning methods, this learning-based method fully analyzes region features of hypotheses for better performance, effectively avoiding differentiation through sampling and solving processes.

	\item LNR introduces a coarse-level hypotheses generation module. It leverages random minimum set sampling for robustness and employs confidence-based selection to prioritize good minimum sets for coarse-hypothesis generation. Furthermore, a fine-level module encodes region features of hypotheses to handle noise and model overlap, which are then used for hypotheses refinement and assessment.
 
	\item Extensive evaluations across four classic robust multi-model fitting tasks show that LNR achieves state-of-the-art performance in multi-line fitting, vanishing point estimation, two-view plane segmentation, and two-view motion tasks. Analysis indicates broad applicability of LNR to various robust multi-model fitting tasks.
    
\end{itemize}

\begin{figure*}[t]
  \centering
   \includegraphics[width=0.99\linewidth]{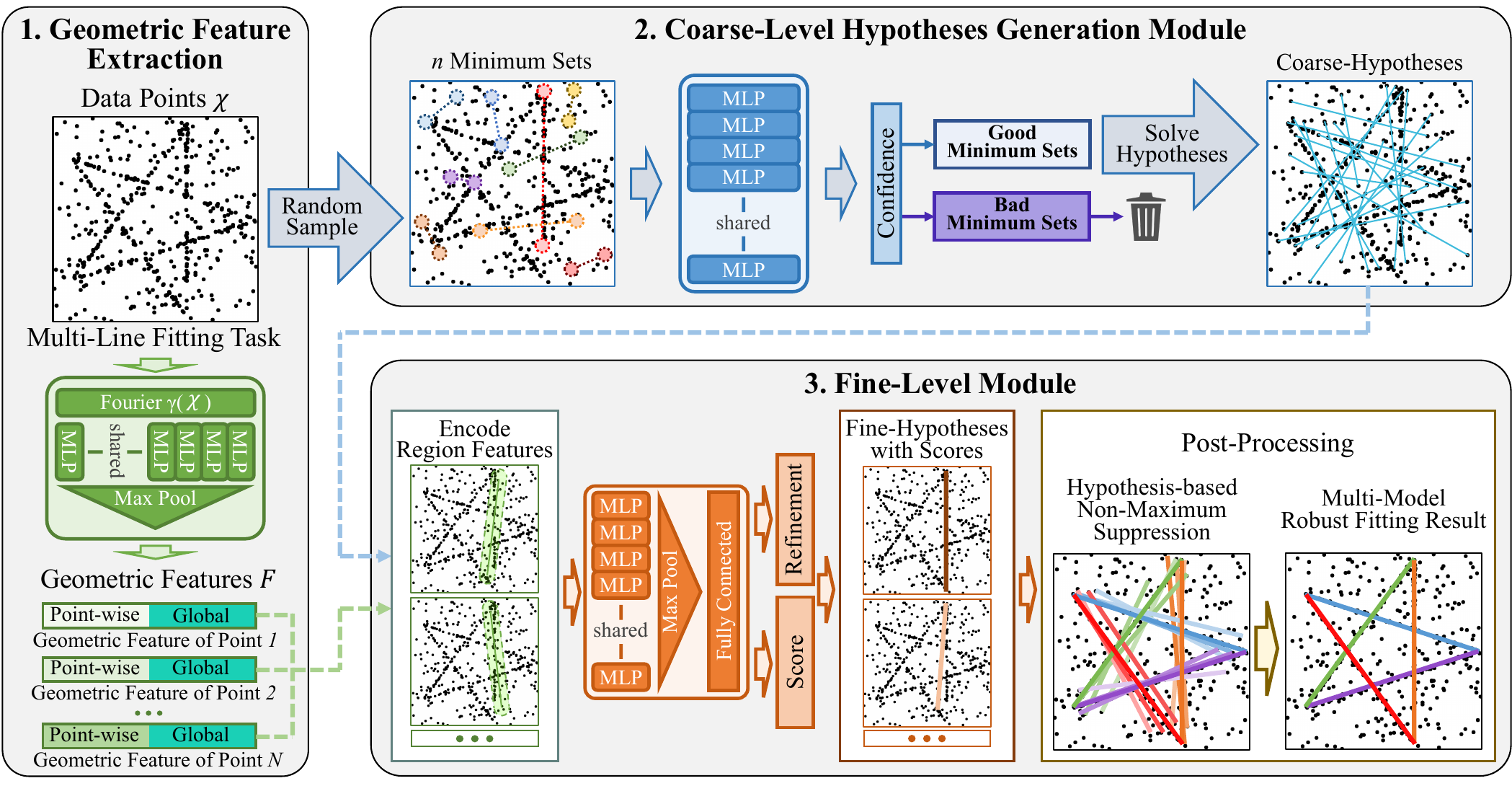}
    \vspace{-10pt}
   \caption{\textbf{The pipeline of the proposed LNR.} Using the multi-line fitting task as an example. Different colored lines indicate their scores and categories. Two points can form a minimum set to determine a line. LNR consists of three main components: \textbf{1.} Geometric features are extracted from the data points (Sec. \ref{geometric}). \textbf{2.} The quality of randomly sampled minimum sets is evaluated. The good minimum sets are solved into coarse-hypotheses (Sec. \ref{coarse}). \textbf{3.} For each coarse-hypotheses, encoding its region feature, which is used for refinement and assessment to get fine-hypotheses (Sec. \ref{refine}). Then a Hypothesis-based Non-Maximum Suppression (HNMS) outputs a compound model as the final robust multi-model fitting results (Sec. \ref{post}).}
   
   \vspace{-10pt}
   \label{fig:pipeline}
\end{figure*}

\section{Related Work}
\vspace{-4pt}
\label{sec:Related_Work}
\subsection{Extension of Single-Model Robust Estimation}
\vspace{-4pt}
Robust multi-model fitting attracts extensive exploration due to its significance. Traditional single-model estimation tasks are often addressed using RANSAC-based methods \cite{chum2003locally, torr2002napsac}. Given the success of these single-model approaches, researchers have naturally sought to extend them to the more complex scenario of multi-model fitting. For instance, Sequential RANSAC \cite{vincent2001detecting} iteratively detects models by clustering inliers, preserving the robustness and generalization of RANSAC. In contrast, MultiRANSAC \cite{zuliani2005multiransac} attempts to fit multiple models by sampling $\mathcal{K}$ models simultaneously to avoid data reuse issues. However, this approach requires $\mathcal{K}$ as a prior knowledge and faces computational inefficiency as $\mathcal{K}$ grows, since finding multiple good models in one iteration is challenging.

\vspace{-10pt}
\subsection{Two Steps Frameworks}
\vspace{-4pt}
Nowadays, most methods \cite{purkait2016clustering, barath2019progressive, amayo2018geometric, barath2018multi} often adopt a two-step framework: hypothesis generation followed by instance selection.
First, a large number of minimum sets are sampled. Then, these minimum sets will generate a hypothesis pool, where the hypotheses are filtered by various numerical optimization methods. Second, many methods select multiple instances from the hypothesis pool to form a compound model. Following the idea of RANSAC, the selected compound model is generally required to have the most inliers.

In the hypothesis pool generation step, some methods aim to improve efficiency by generating only necessary hypotheses. For example, Chin \textit{et al.} \cite{chin2011accelerated} employ residual ordering information to guide the sampling process. Purkait \textit{et al.} \cite{purkait2016clustering} propose a hyperedge-guided sampling strategy based on the concept of clustering.
MF-Net \cite{barath2022learning} directly classifies the minimum sets as valid or invalid through a neural network. T-Linkage \cite{magri2014t} performs clustering via preference sets. However, these clustering-style techniques ignore the reuse of data.
Prog-X \cite{barath2019progressive} removes bad instances through the design of proposal validation and optimizing modules to improve efficiency. CONSAC extends the NG-RANSAC \cite{brachmann2019neural} method from single-model robust estimation to multi-model fitting tasks. Specifically, CONSAC simultaneously feeds generated hypotheses information along with data points into a network. Then the network produces the probabilities of data points, which can guide the sampling for this iteration. PARSAC \cite{kluger2024parsac} improves its efficiency by parallel computing. But the number of instances is needed as a priori information, which limits the application of PARSAC.

In the selecting instances step, several approaches aim to evaluate the qualities of hypotheses \cite{isack2012energy, amayo2018geometric, barath2018multi}.
PEARL \cite{isack2012energy} addresses geometric multi-model fitting by viewing it as an optimization problem. This approach employs a global energy function that considers both geometric errors and the intrinsic regularity of clusters. Amayo \textit{et al.} \cite{amayo2018geometric} explore the soft assignments of points to the geometric model by minimizing the overall assignment energy. In addition, there are methods \cite{wang2015mode} that treat instances as vertices and points as hyper-edges, solving the evaluation problem through hyper-graph partitioning. MSH \cite{wang2015mode} regards the multi-model fitting task as a mode-seeking problem on a hyper-graph.

Building on the above methods, we further remodel the two-step framework into a deep learning framework as LNR. It extracts geometric features and selects good minimum sets in the coarse-level to improve efficiency. At the fine-level, LNR encodes region features for scoring and refining these hypotheses, avoiding differentiating the hypotheses and addressing model overlap.

\vspace{-7pt}
\section{Method}
\vspace{-4pt}
We design a coarse-to-fine framework for robust multi-model fitting. Fig. \ref{fig:pipeline} provides an overview of our proposed method, which comprises three key stages: geometric feature extraction (Sec. \ref{geometric}), coarse hypothesis generation (Sec. \ref{coarse}), and fine-level refinement (Sec. \ref{coarse}).

\subsection{Geometric Feature Extraction}
\label{geometric}
\begin{figure}[t]
  \centering
   \includegraphics[width=1.0\linewidth]{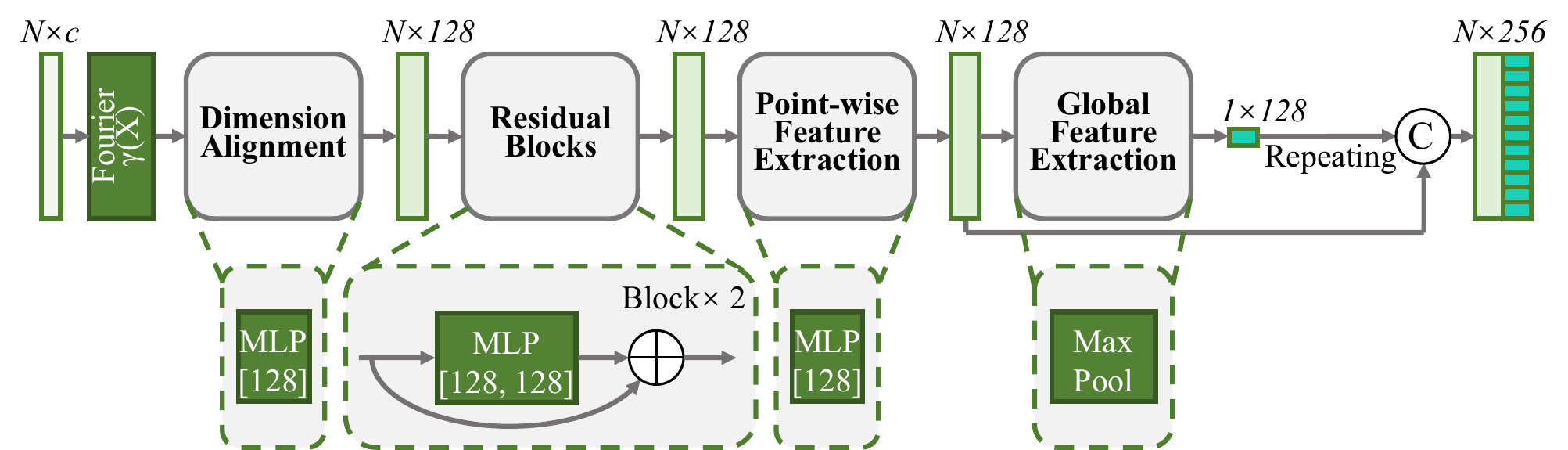}
     \vspace{-20pt}
    \caption{\textbf{The architecture of the neural network in geometric feature extraction.} Most raw data can be represented as unordered $N \times c$ data points, making the input data $N \times c$. The symbol $\oplus$ denotes element-wise vector addition, while ``C" denotes vector concatenation. The network output is 256-dimensional, comprising 128-dimensional point-wise features concatenated with 128-dimensional global features.}
   \label{fig:GFE}
   \vspace{-0.5cm}
\end{figure}

For a given set of input data points, denoted as $\chi = \left\{ {{{\rm{x}}_i}} \right\}_{i = 1}^N$ contaminated by noise, we first extract geometric features $F$ that capture both local and global information. As shown in Fig. \ref{fig:GFE}, raw data points undergo Fourier feature mapping \cite{tancik2020fourier} to enhance spatial frequency encoding. Shared multilayer perceptrons (MLPs) (implemented via $1 \times 1$ convolutions) then generate point-wise features for each data point. Subsequently, a max-pooling operation aggregates these point-wise features to generate global features, representing the holistic properties of the input data. These point-wise and global features are concatenated to form a 256-dimensional geometric feature vector $F$ for each point. This combined representation enables subsequent modules to leverage both local point-specific details and global contextual information.

\subsection{Coarse-level Hypotheses Generation Module}
\label{coarse}
After extracting geometric features, LNR generates coarse-hypotheses through random sampling akin to RANSAC. Specifically, LNR draws $n$ minimum sets $\mathbb{M}  = \left\{ {{M_j}} \right\}_{j = 1}^n$ from $\chi$, where each set $M_j$ contains the minimum number of points required for model estimation (e.g., two points for line fitting). The number of minimum sets, $n$, is intentionally large to ensure a comprehensive exploration of potential hypotheses.

\begin{figure}[!t]
  \centering
   \includegraphics[width=1.0\linewidth]{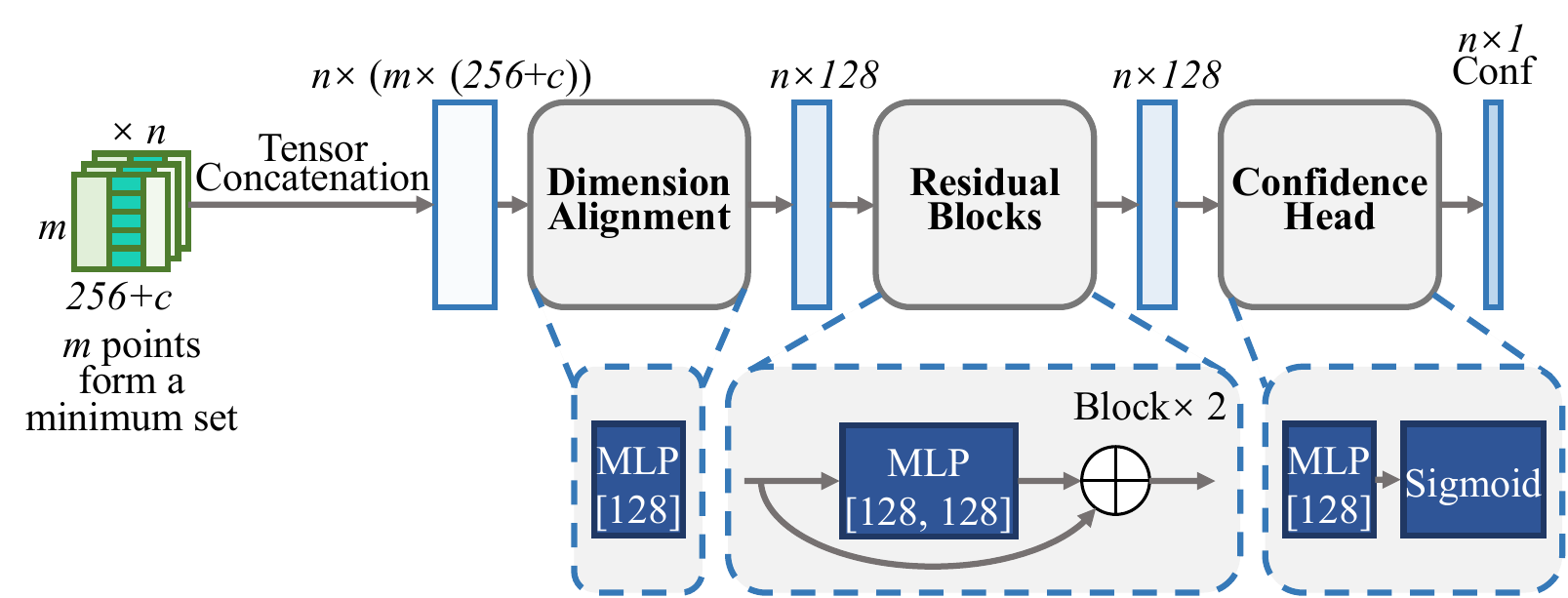}
    \vspace{-20pt}
   \caption{\textbf{The architecture of the neural network in coarse-level module.} The input data consists of the features of each point in the $n$ minimum sets. The size $m$ of the minimum set varies with the task. The features of each point are concatenated with $c$-dimensional original features and the extracted 256-dimensional geometric features. The tensor concatenation module concatenates the features of all the points of a minimum set.}
   \label{fig:ER}
   \vspace{-10pt}
\end{figure}

Among the sampled minimum sets $\mathbb{M}$, a large number of sets are incorrect. To select a small number of good minimum sets to improve efficiency, avoiding numerous of bad minimum sets consuming resources, LNR introduces a coarse-level module to estimate confidence of each sampled minimum set. This module uses a neural network, illustrated in Fig. \ref{fig:ER}, to fully analyzes the geometric features of the minimum sets to assign confidence. Specifically, the geometric features of the points within a minimum set, $F_{M}$, are concatenated to form a minimum set feature. To preserve original data information, the original features are also incorporated at each point. Features from all minimum sets, $\mathbb{F}_{M}$, are then processed together. Then, the LNR analyzes these features using shared MLPs and generates a confidence $\mathcal{F}_{conf}(M_j)$ for each minimum set. A higher confidence indicates a minimum set of better quality. Only the top $n_{good}$ minimum sets, $\mathbb{M}_{good}$, with the highest confidence scores are then processed by a minimum solver $S$. This solver produces coarse-hypotheses $H_{coarse} = \left\{ {S\left( {{M_j}} \right)\left| {{{M_j}} \in \mathbb{M}_{good},j = 1,2,...,n_{good}} \right.} \right\}$. Thus, even though a large number of minimum sets are initially sampled, only a small fraction of the minimum sets will be solved, greatly improving the overall efficiency.

\subsection{Fine-level Module}
\label{refine}

To refine and assess each coarse-hypothesis, LNR employs a fine-level module. This module prioritizes local information over global context, as local details are more relevant for refining individual hypotheses \cite{fischler1981random, barath2020magsac++, chum2003locally}. In contrast, while global information is comprehensive, it can introduce many irrelevant details that may hinder the refinement process for a specific hypothesis.

\begin{figure}[t]
  \centering
   \includegraphics[width=0.75\linewidth]{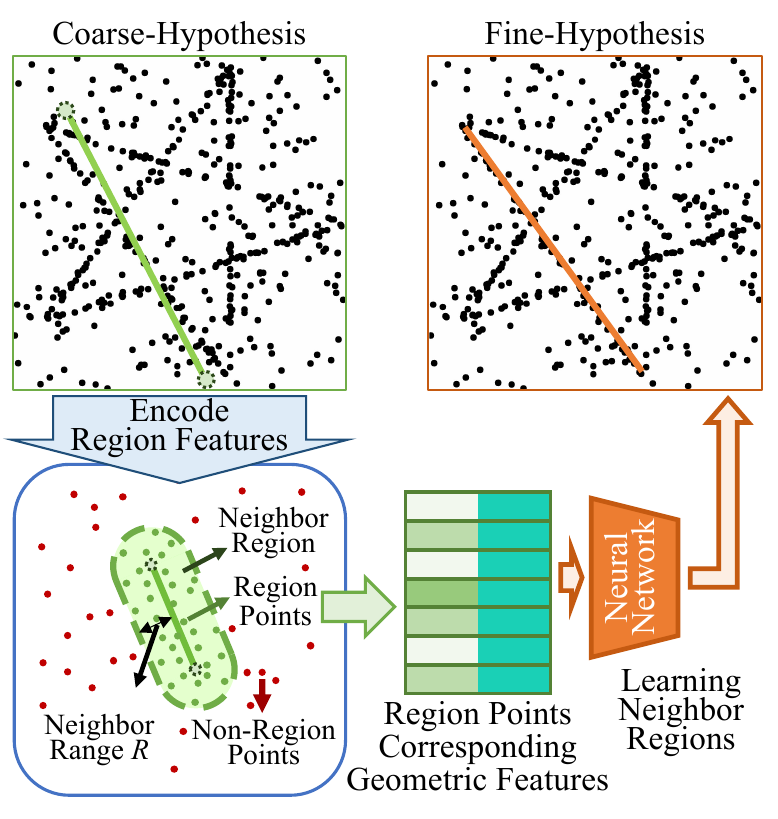}
    \vspace{-10pt}
   \caption{\textbf{Encoding region features in fine-level module.} Each coarse-hypothesis is encoded region features. The data points within a distance $R$ of the hypothesis are delimited as region points. Geometric features associated with these region points are concatenated as region features, which are learned by the neural network for refinement. Learning the region features of the hypothesis rather than the hypothesis parameters avoids differentiating the sampling process and the model solvers.}
   \vspace{-10pt}
   \label{fig:region}
\end{figure}

As depicted in Fig. \ref{fig:region}, the fine-level module begins by encoding a region feature for each hypothesis. For each hypothesis, a neighbor region is defined as the area within a range $R$ in the data point space. The definition of this range $R$ and the distance metric used vary depending on the task. For example, in multi-line fitting, it is a Euclidean distance; for vanishing point estimation, it is a directional distance; for two-view plane segmentation, it is re-projection error; and for two-view motion, it is normalized epipolar constraint distance. The geometric features $F$ of the data points within these neighbor regions are then encoded as the region feature $F_{region}$. Since the number of points within each region can vary, regions with fewer points are padded with zeros to ensure consistent input size for a data batch. Inspired by MQ-Net \cite{barath2022learning}, LNR concatenates residuals from each point in the neighbor region to the hypothesis. 

These enriched features are then fed into a network with a refinement head and a score head. The refinement head predicts increments to the hypothesis parameters. These increments are added to the coarse-hypotheses $H_{coarse}$ to yield the refined fine-hypotheses $H_{fine}$. The score head, with a single output channel, assesses the quality of each hypothesis, providing a more robust evaluation than simple inlier counting (verified in the ablation study (Sec. \ref{ablation}). This fine-level module thus refines the initial coarse-hypotheses by focusing on local region information and learning to score and adjust hypothesis parameters.

\begin{figure}[t]
  \centering
   \includegraphics[width=1.0\linewidth]{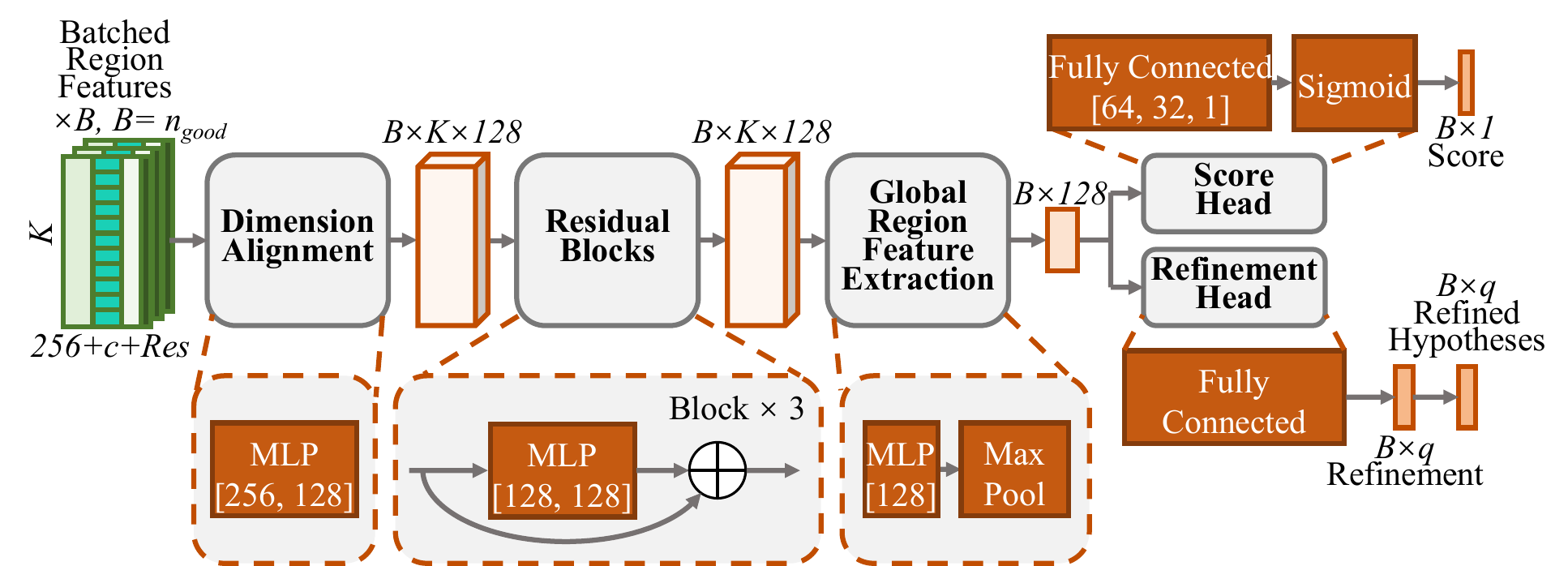}
    \vspace{-20pt}
    \caption{\textbf{The architecture of the neural network in the fine-level module.} The input data consists of the encoded features of the neighbor region points, forming a batch. Based on the features input to the coarse-level module, the residuals from each point in the neighbor region to its hypothesis are also concatenated. The symbol $\oplus$ denotes element-wise vector addition. The Refinement Head outputs only the increments of the hypothesis parameters, which need to be added to the original hypotheses to obtain the refined hypotheses.}
   \label{fig:RSR}
   \vspace{-10pt}
\end{figure}

\subsection{Testing Post-Processing and Training Loss}
\label{post}
\begin{figure}[!t]
  \centering
   \includegraphics[width=0.2\textwidth]{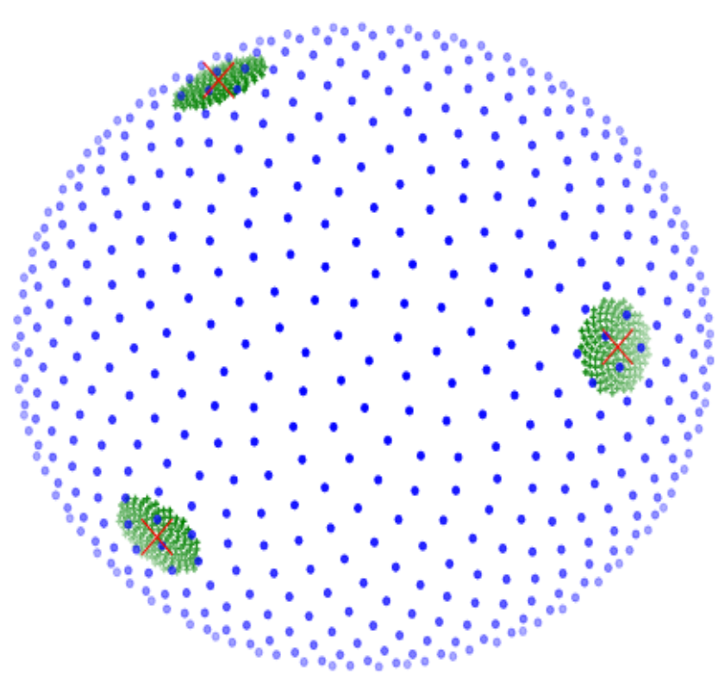}
    \vspace{-10pt}
   \caption{\textbf{The similarity between vanishing points.} The vanishing points are projected onto a Gaussian sphere and the distances on the sphere are used to compute the similarity between the vanishing points.} 
   \label{fig:supp_vp_similarity}
\end{figure}

During testing, the goal is to obtain a set of non-redundant, high-quality hypotheses that represent the underlying multi-model structure. To achieve this, LNR employs Hypothesis-based Non-Maximum Suppression (HNMS) to filter the fine-hypotheses $H_{fine}$. HNMS, inspired by the method of Lin \textit{et al.} \cite{lin2022deep} and Non-Maximum Suppression in object detection \cite{redmon2016you}, iteratively selects the best hypotheses based on their fine-level scores and suppresses geometrically similar and redundant hypotheses. For the calculation of similarity between hypotheses, (1) the multi-line fitting task uses the distance between two endpoints and the angle between the two lines to compute the similarity. (2) For the vanishing point estimation task, the VPs can be projected onto the Gaussian sphere \cite{lin2022deep}, where proximity on the sphere indicates similarity, as shown in Fig. \ref{fig:supp_vp_similarity}. (3) In the two-view plane segmentation and two-view motion tasks, which are ``point-to-model assignment" problems, the distance from each point to the hypothesis is encoded as a vector. The similarity between homographies and fundamental matrices is then computed based on the similarity between these vectors. The final set of selected hypotheses from HNMS constitutes the multi-model fitting result.

\begin{figure}[t]
  \centering
   \includegraphics[width=0.80\linewidth]{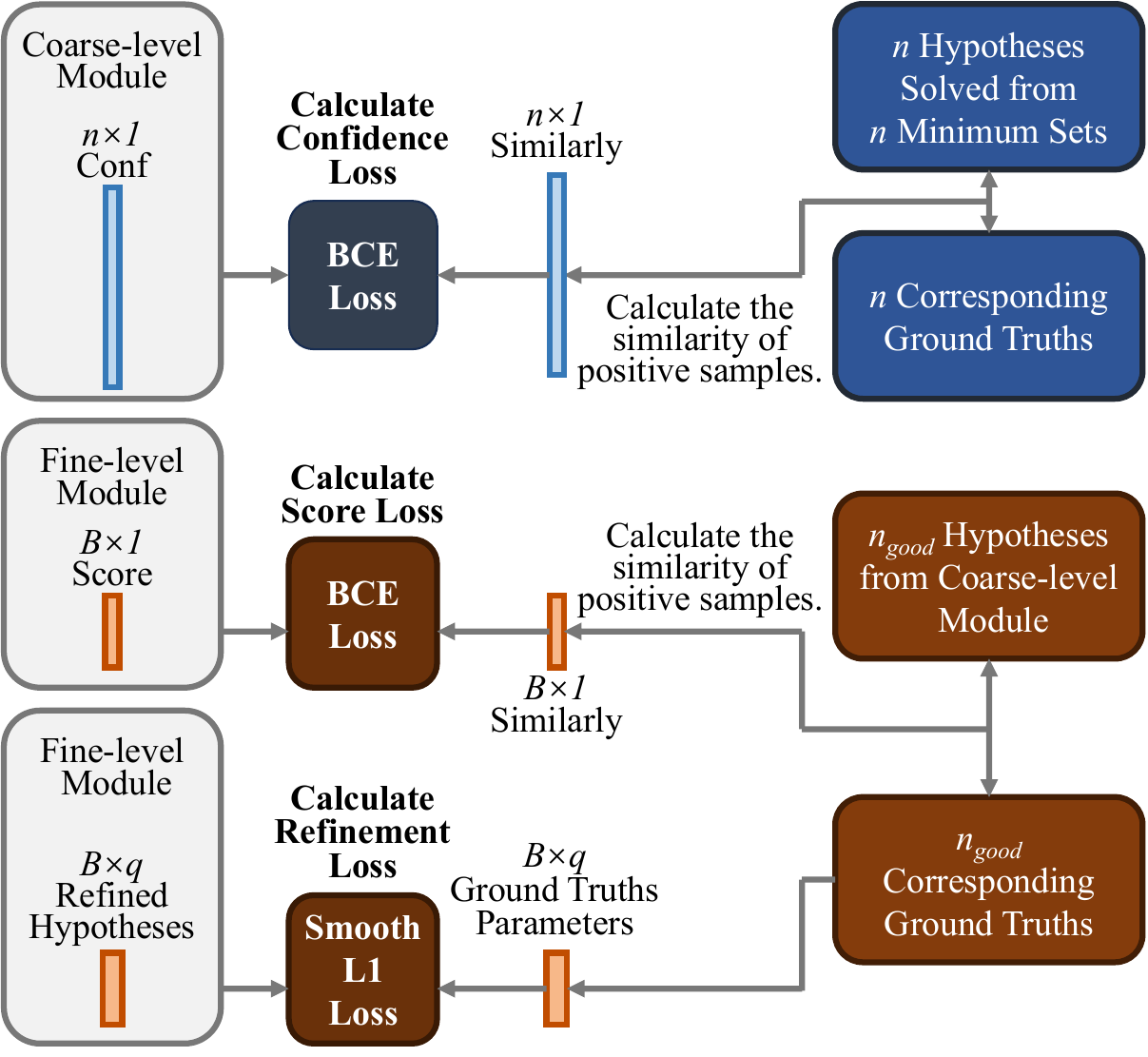}
    \vspace{-10pt}
   \caption{\textbf{The architecture of the neural network in fine-level module.} The LNR optimizes three loss functions: Confidence Loss, Score Loss, and Refinement Loss. Both Confidence Loss and Score Loss utilize Binary Cross Entropy Loss (BCE Loss). The coarse-level module identifies the most similar ground truths for loss after solving the hypotheses for the minimum sets of all sampled minimum sets. The fine-level module refines coarse-hypotheses by finding the most similar ground truths for the coarse-hypotheses. To enhance training efficiency, BCE Loss is applied only to positive samples. The Refinement Loss uses Smooth L1 Loss for regression.}
   \label{fig:loss}
   \vspace{-5pt}
\end{figure}

The training of LNR involves optimizing three loss functions: Confidence Loss ($\mathcal{L}_{Conf}$), Score Loss ($\mathcal{L}_{Score}$), and Refinement Loss ($\mathcal{L}_{Refine}$), as illustrated in Fig. \ref{fig:loss}. The Confidence Loss, applied in the coarse-level module, and the Score Loss, applied in the fine-level module, both use Binary Cross Entropy (BCE) loss for binary classification. For Confidence Loss, minimum sets are labeled as ``good" (1) or ``bad" (0) based on the similarity of their solved hypotheses to ground truth instances ${y_{con{f_j}}} \in \{ 0,1\}$. This allows the coarse-level module to learn to select promising minimum sets without relying on explicit degeneracy tests. Therefore, Binary Cross Entropy (BCE) loss is used:
\begin{equation}
\begin{split}
{{\mathcal{L}}_{Conf}} & =  - \frac{1}{n}\sum\limits_{j = 1}^n {[{y_{con{f_j}}}}  \cdot \log \left( {{{\cal F}_{conf}}({M_j})} \right)\\
 & + \left( {1 - {y_{con{f_j}}}} \right) \cdot \log \left( {1 - {{\cal F}_{conf}}({M_j})} \right)].
\end{split}
\end{equation}

For the Score Loss, the label ${y_{scor{e_j}}}$ also comes from the similarity between the hypothesis $h_j$ and the ground truth instance $I_{gt}$. The difference is that the confidence of the minimum sets $\mathbb{M}$ is predicted in the coarse-level module. While the score of the hypothesis $h_j$ is predicted in the fine-level module.
Therefore, the scoring part is also a binary classification problem with BCE loss:
\begin{equation}
\begin{split}
{{\mathcal{L}}_{score}} & =  - \frac{1}{{{n_{good}}}}\sum\limits_{j = 1}^{{n_{good}}} {[{y_{scor{e_j}}}}  \cdot \log \left( {{{\cal F}_{score}}({h_j})} \right)\\
 & + \left( {1 - {y_{scor{e_j}}}} \right) \cdot \log \left( {1 - {{\cal F}_{score}}({h_j})} \right)].
\end{split}
\end{equation}

The Refinement Loss utilizes Smooth L1 loss to regress the offsets needed to refine the hypothesis parameters. For the multi-line fitting task, the predicted refinement offsets for the two endpoints of the hypothetical line segment are denoted as $\Delta {x_1},\Delta {y_1},\Delta {x_2},\Delta {y_2}$. These offsets are then applied to the coordinate ${x_{{1_h}}}, {y_{{1_h}}}, {x_{{2_h}}}, {y_{{2_h}}}$, forming the regression residuals between ground truth and hypothesis:
\begin{equation}
\begin{array}{l}
{r_{{x_1}}} = {x_{{1_{gt}}}} - ({x_{{1_h}}} + \Delta {x_1}),{r_{{y_1}}} = {y_{{1_{gt}}}} - ({y_{{1_h}}} + \Delta {y_1}),\\
{r_{{x_2}}} = {x_{{2_{gt}}}} - ({x_{{2_h}}} + \Delta {x_2}),{r_{{y_2}}} = {y_{{2_{gt}}}} - ({y_{{2_h}}} + \Delta {y_2}).
\end{array}
\end{equation}
The smooth L1 \cite{girshick2015fast} regression loss can represent this regression:
\begin{equation}
{{\cal L}_{Refine}} = \sum\limits_{j = 1}^{{n_{good}}} {\sum\limits_{q \in ({x_1},{y_1},{x_2},{y_2})} {{{\rm{Smooth}}L1}\left( {{r_q}} \right)} }.
\end{equation}

The regression residuals of vanishing point estimation is
\begin{equation}
\begin{array}{l}
{r_{{x}}} = {x_{{{gt}}}} - ({x_{{h}}} + \Delta {x}),\\{r_{{y}}} = {y_{{{gt}}}} - ({y_{{h}}} + \Delta {y}),\\
{r_{{z}}} = {z_{{{gt}}}} - ({z_{{h}}} + \Delta {z}).
\end{array}
\end{equation}
The smooth L1 regression loss can represent this regression:
\begin{equation}
{{\cal L}_{Refine}} = \sum\limits_{j = 1}^{{n_{good}}} {\sum\limits_{q \in ({x},{y},{z})} {{{\rm{Smooth}}L1}\left( {{r_q}} \right)} }.
\end{equation}

For the two-view plane segmentation task, the predicted offset is $\Delta {H}$. Then the regression residual is:
\begin{equation}
\begin{array}{l}
{r_{{H}}} = {H_{{{gt}}}} - ({H_{{h}}} + \Delta {H}).
\end{array}
\end{equation}
Thus the smooth L1 regression loss of this task can represent:
\begin{equation}
{{\cal L}_{Refine}} = \sum\limits_{j = 1}^{{n_{good}}} {{{\rm{Smooth}}L1}\left( {{r_H}} \right) }.
\end{equation}

For the two-view motion task, it is similar to the two-view plane segmentation task. The predicted offset is $\Delta {F}$. Then the regression residual is:
\begin{equation}
\begin{array}{l}
{r_{{F}}} = {F_{{{gt}}}} - ({F_{{h}}} + \Delta {F}).
\end{array}
\end{equation}
Thus the smooth L1 regression loss of this task is:
\begin{equation}
{{\cal L}_{Refine}} = \sum\limits_{j = 1}^{{n_{good}}} {{{\rm{Smooth}}L1}\left( {{r_F}} \right) }.
\end{equation}

Thus, the total loss function is:
\begin{equation}
\begin{split}
{{\cal L}_{Total}} = {{\cal L}_{Conf}} + {{\cal L}_{Score}} + {{\cal L}_{Refine}}.
\label{totalloss}
\end{split}
\end{equation}
In this way, this multi-task loss can jointly optimizes all modules towards a common goal, improving overall performance. Moreover, the training of this multi-task loss function allows the loss of the fine-level module to optimize coarse-level module as well. This means that the coarse-level module produces coarse-hypotheses $H_{coarse}$ that are more consistent with the requirements of the fine-level module.

\section{Experiments}
\label{Experiments}

This section details the experiments conducted to evaluate the performance of LNR across four distinct and classic computer vision tasks: multi-line fitting, vanishing point estimation, two-view plane segmentation, and two-view motion estimation. Multi-line fitting serves as a foundational task for robust multi-model fitting algorithms. It offers a clear visualization of an algorithm to various noise scales and its ability to manage model overlap, as it can manipulate the degree of overlap between models. Vanishing point estimation, conversely, showcases algorithm performance in complex, real-world scenarios. Finally, the two-view plane segmentation and two-view motion estimation tasks assess the capability of LNR in processing correspondences between image pairs.  Crucially, these two tasks represent classic instances of ``point assignment to model" problems.

\subsection{Settings}
\label{settings}
The LNR framework undergoes training for 100 epochs. Post-processing is not incorporated during the training phase. All experiments are executed on an RTX 3090 GPU, with the framework implemented using PyTorch. The computational efficiency of LNR is O(n).

\subsection{Case Study 1: Multi-Line Fitting}
This task evaluates basic multi-model fitting capabilities using synthetic 2D point data $\chi = \left\{ {\left[ {{x_i},{y_i}} \right]\left| {\ i = 1,2,... ,N} \right.} \right\} \in \mathbb{R}^{N \times c}$, where $c=2$ indicates that the data points only contain the coordinates. As a line can be defined by two points, each minimum set consists of two data points. For comparison, several classic methods are also evaluated in the same scenes with the same settings.

A synthetic dataset is generated to systematically evaluate performance, encompassing scenes with 1 to 10 lines, each scene containing 12,000 pictures. Of these, 10,000 pictures are designated for training and the remaining 2,000 for testing. All data points are constrained within a $1 \times 1$ picture. Lines, ranging from 1 to 10 per picture, are randomly generated with a minimum length of 0.3 times the longest distance of the picture. Each line segment comprises 40 to 100 points, randomly distributed and perturbed by Gaussian noise $\mathcal{N}(0,{\sigma ^2})$, with $\sigma$ varying between 0.007 and 0.008. Outliers, constituting 40\% to 60\% of the total data points, are added uniformly and randomly. Note that outliers close to line segments may become inliers due to random scattering.

Performance assessment relies on the Area Under the Curve (AUC) metric, as detailed in Kluger \textit{et al.} \cite{kluger2020consac}. AUC calculation generating a recall curve for all line segments in the test set, considering a maximum angular error of $0.5^{\circ}$. AUC can comprehensively assess the accuracy and recall of a method.

TABLE \ref{tab:line_result} summarizes the quantitative performance of LNR and comparative methods across scenes with increasing line counts. LNR consistently outperforms other methods across all scenarios, demonstrating its effective feature learning for accurate instance generation. As the number of lines increases, a general performance decrease is observed across all methods, attributable to heightened model overlap and noise. However, the performance of alternative methods degrades significantly, highlighting their limitations in managing model overlap. Conversely, our proposed method exhibits a minimal performance reduction, underscoring its robust resistance to interference and effective handling of model overlap, properties attributed to its effective region feature encoding.

\begin{table}[t]\scriptsize 
\centering
\caption{\textbf{The quantitative result of multi-line fitting.} The average AUC@$0.5^{\circ}$ ↑ on 1-10 line scenarios with five runs are reported. The best indicators are bolded.}
\vspace{-5pt}
\setlength{\tabcolsep}{0.33mm}
\renewcommand\arraystretch{1.0}
\begin{tabular}{l||cccccccccc}
\toprule
Method             & 1               & 2               & 3               & 4               & 5               & 6               & 7               & 8               & 9               & 10              \\ \hline \hline
Seq. RANSAC \cite{vincent2001detecting} & 91.98          & 65.76          & 58.16          & 53.80          & 51.95          & 50.66          & 50.01          & 49.28          & 49.02          & 48.55          \\
J-linkage \cite{toldo2008robust} & 93.14          & 74.98          & 67.98          & 66.87          & 66.02          & 65.68          & 65.17          & 64.97          & 64.63         & 63.97          \\
T-linkage \cite{magri2014t} & 94.31          & 77.90          & 69.97          & 68.01          & 67.18          & 66.61          & 66.10          & 65.71          & 65.10          & 64.29          \\
CONSAC \cite{kluger2020consac} & 96.14          & 91.29          & 86.63          & 84.21          & 81.61          & 79.83          & 78.30          & 76.34          & 75.51          & 74.84          \\
\textbf{Ours}               & \textbf{96.73} & \textbf{93.41} & \textbf{89.99} & \textbf{87.71} & \textbf{84.89} & \textbf{83.32} & \textbf{81.47} & \textbf{79.91} & \textbf{78.59} & \textbf{77.51} \\ \bottomrule
\end{tabular}
\label{tab:line_result}
\vspace{-8pt}
\end{table}

\begin{figure}[t]
  \centering
   \includegraphics[width=0.8\linewidth]{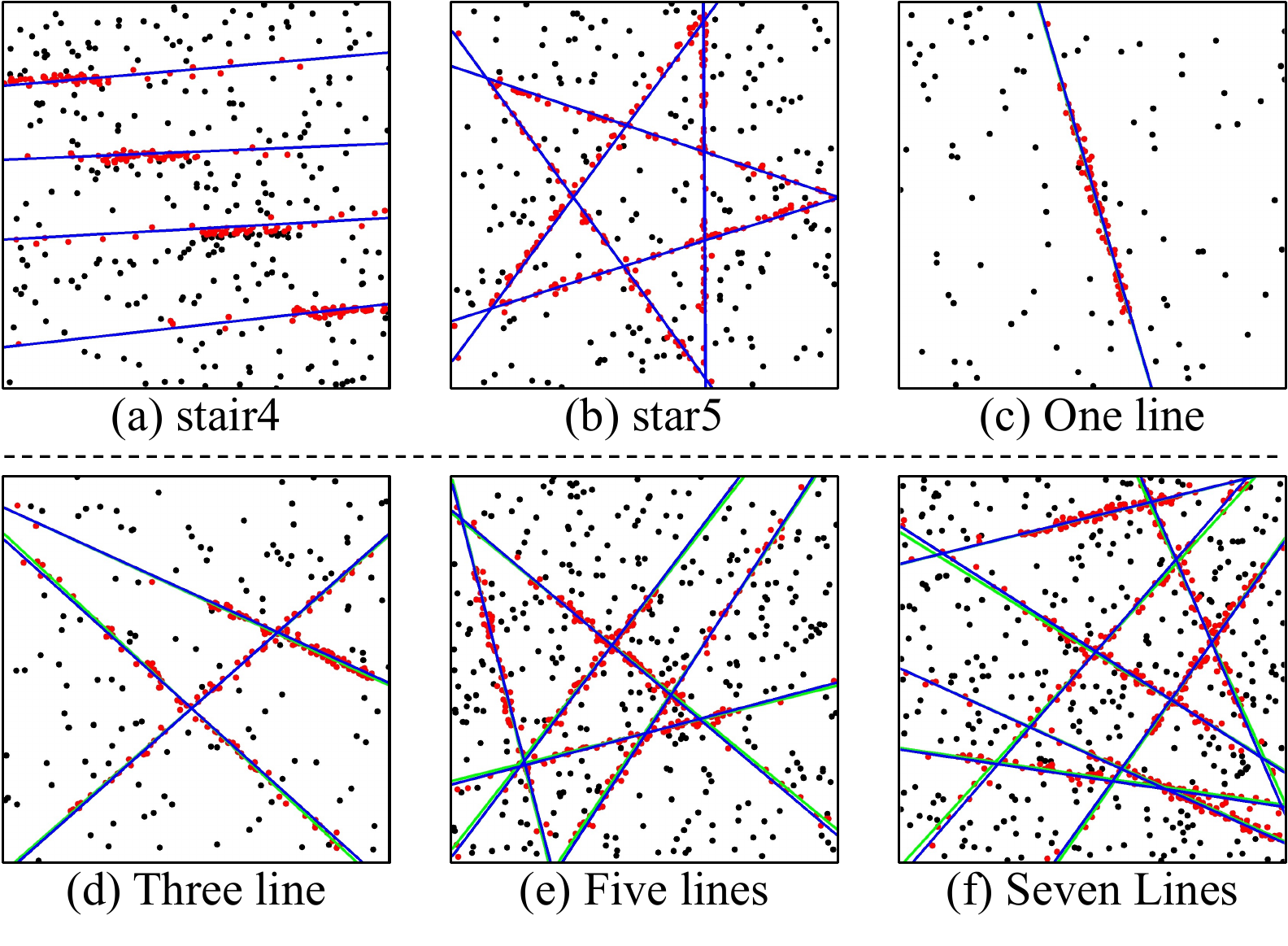}
    \vspace{-15pt}
   \caption{\textbf{The qualitative results of LNR on multi-line fitting.} The red points represent inliers, while the black points are outliers. The ground truth is represented by green lines, while the fitted lines are represented by blue lines.} 
   \vspace{-8pt}
   \label{fig:lines}
\end{figure}

Qualitative evaluations, shown in Fig. \ref{fig:lines} (a) and (b) using an existing dataset \cite{toldo2008robust}, illustrate that our proposed method can accurately fit lines. Moreover, Fig. \ref{fig:lines} (c) demonstrates the efficacy of our proposed method in fitting a single line, indicating its capability in single-model robust estimation tasks. Furthermore, even with increased line counts and noise levels, as shown in Fig. \ref{fig:lines} (d), (e), and (f), LNR maintains accurate multi-instance fitting. These qualitative and quantitative results collectively suggest that our proposed method effectively leverages region features to address the challenges posed by model overlap.

\begin{table*}[t]\footnotesize 
\centering
\caption{\textbf{The quantitative result of vanishing point estimation task.} The average AUC in \% and related standard deviations (std.) over five runs are reported. The general robust fitting methods are compared. The ``General" is ``Yes" for a general robust fitting method, and ``No" for a vanishing point-specific method. The ``DL" is ``Yes" for a deep learning method. The best indicators are bolded and sub-optimal indicators are underlined.}
\vspace{-5pt}
\setlength{\tabcolsep}{1.7mm}
\renewcommand\arraystretch{1.0}
\begin{tabular}{l||c|c|cccc|cccc|cccc|c}
\toprule
\multirow{2}{*}{Method} & \multirow{2}{*}{General} & \multirow{2}{*}{DL}& \multicolumn{4}{c|}{NYU-VP}    & \multicolumn{4}{c|}{YUD+}      & \multicolumn{4}{c|}{YUD}       & \multirow{2}{*}{\begin{tabular}[c]{@{}c@{}}Speed\\ (Hz)↑\end{tabular}} \\ \cline{4-15}
                        &              &       & @$5^{\circ}$↑ & std.↓ & @$10^{\circ}$↑ & std.↓ & @$5^{\circ}$↑ & std.↓ & @$10^{\circ}$↑ & std.↓ & @$5^{\circ}$↑ & std.↓ & @$10^{\circ}$↑ & std.↓ &                            \\ \hline\hline
J-Linkage \cite{toldo2008robust}               & Yes     & No             & 42.22  & 0.82 & 56.60   & 0.83 & 60.42  & 1.16 & 72.39   & 0.85 & 68.71  & 2.26 & 81.10   & 1.55 & 0.52                       \\
T-Linkage \cite{magri2014t}              & Yes   & No                  & 43.03  & 0.71 & 57.79   & 0.72 & 59.45  & 0.84 & 71.75   & 0.55 & 66.11  & 1.43 & 79.54   & 0.92 & 1.28                       \\
Prog-X \cite{barath2019progressive}           & Yes     & No                  & 49.28  & 0.20 & 60.73   & 0.22 & 60.03  & 0.71 & 68.54   & 0.70 & 60.11  & 0.78 & 68.48   & 0.74 & 29.42                      \\
CONSAC \cite{kluger2020consac}             & Yes    & Yes                 & 50.61  & 0.39 & 64.29   & 0.40 & 63.76  & 0.68 & 74.45   & 0.76 & 71.34  & 0.48 & 82.81   & 0.30 & 0.31                       \\
Lin \textit{et al.} \cite{lin2022deep}               & No   & Yes                       & \underline{55.87}       & -                 & \underline{69.53}      & -                  & 59.57   & -                     & 71.34        & -      & 72.25  & -     & 84.98        & -          & 5.50                        \\
PARSAC \cite{kluger2024parsac}             & Yes   & Yes                 & 51.68  & 0.12 & 64.61   & 0.13 & 65.49  & 0.35 & 74.76   & 0.42 & \underline{75.92}  & 0.15 & 86.39   & \underline{0.09} & \textbf{84.13}                      \\
\textbf{Ours}            & Yes      & Yes                 & \textbf{57.54}  & \textbf{0.08} & \textbf{71.93}   & \textbf{0.10} & \textbf{67.96}  & \textbf{0.28} & \textbf{77.21}   & \textbf{0.32} & \textbf{76.39}  & \textbf{0.13} & \textbf{87.12}   & \textbf{0.08} & 10.31                      \\
\textbf{Ours-tiny}        & Yes   & Yes                 & 51.52  & \underline{0.11} & 64.46   & \underline{0.12} & \underline{65.53}  & \underline{0.31} & \underline{74.91}   & \underline{0.34} & \underline{75.92}  & \underline{0.14} & \underline{86.40}   & 0.10 & \underline{46.96}                      \\ \bottomrule
\end{tabular}
\label{tab:vptable}
\vspace{-0.3cm}
\end{table*}

\begin{figure*}[t]
  \centering
   \includegraphics[width=0.99\linewidth]{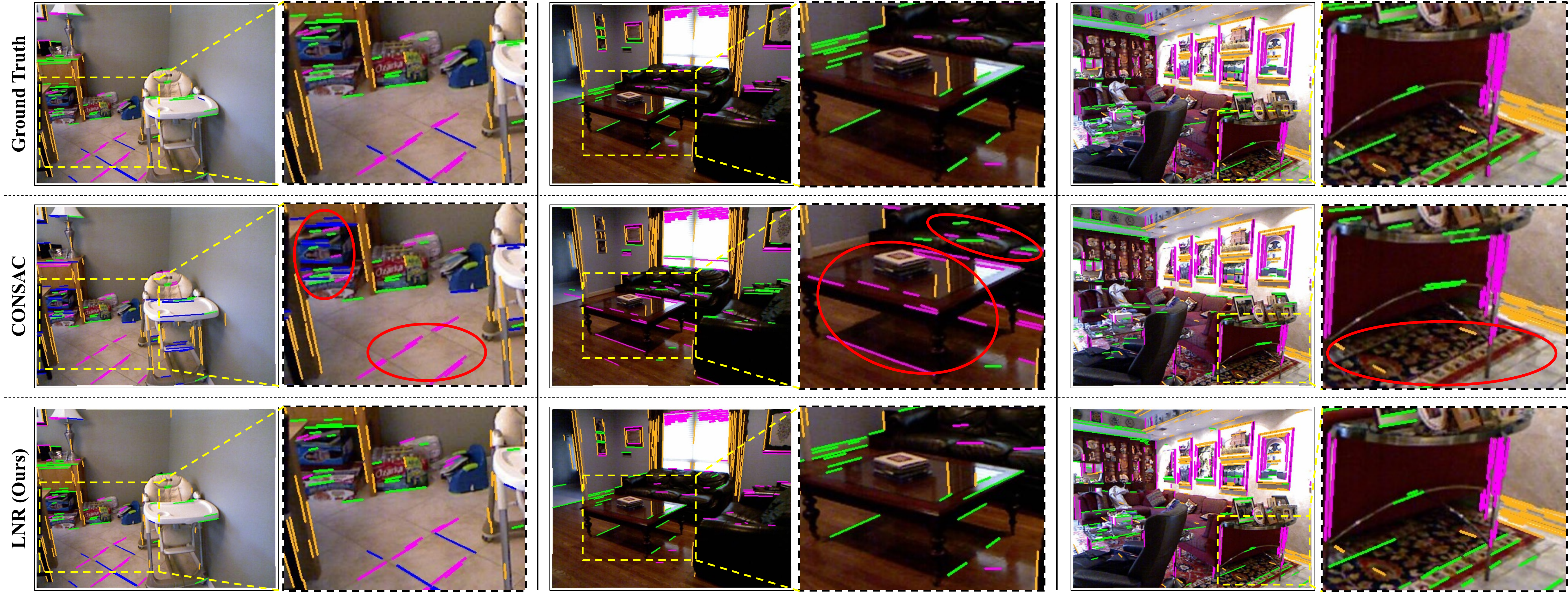}
    \vspace{-10pt}
   \caption{\textbf{The qualitative results of CONSAC and LNR on the vanishing point estimation task in various scenes.} The inlier line segments that belong to different vanishing points are distinguished by different colors. The major errors are indicated by red circles.} 
   \vspace{-10pt}
   \label{fig:vp}
\end{figure*}

\begin{figure*}[t]
  \centering
   \includegraphics[width=0.99\linewidth]{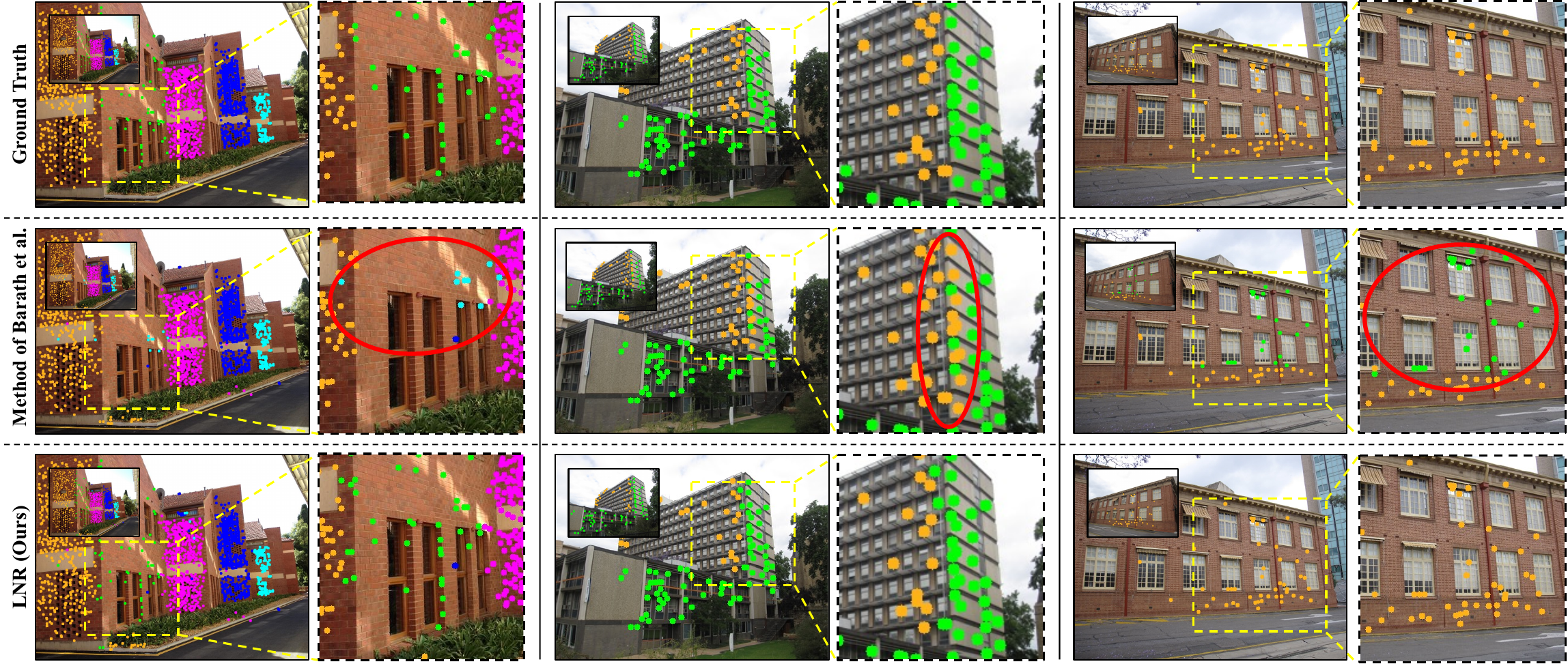}
    \vspace{-10pt}
   \caption{\textbf{The qualitative results of the method of Barath \textit{et al.} \cite{barath2023finding} and LNR on the two-view plane segmentation task in various scenes.} Different colors indicate which of the planes the correspondences belong to. The small image and the large image are a pair of images. Their correspondences are used as input to the task. The major errors are indicated by red circles.} 
   \vspace{-10pt}
   \label{fig:H}
\end{figure*}

\subsection{Case Study 2: Vanishing Point Estimation}
\label{case2}

In this task, data points are represented as $\chi = \left\{ {\left[ {x_1^i,y_1^i,x_2^i,y_2^i} \right]\left| {\;i = 1,2,.... ,N} \right.} \right\}$, where $\chi \in \mathbb{R}^{N \times c}$ and $c=4$, representing the endpoints of line segments. Evaluation settings are consistent with CONSAC \cite{kluger2020consac}. For comparison, several methods are compared. Notably, to test robustness and generalization, LNR is trained exclusively on the NYU-VP dataset and subsequently tested on the YUD+ and YUD datasets without further training.

The quantitative results, presented in TABLE \ref{tab:vptable}, demonstrate strong average AUC performance of LNR. Notably, LNR, as a general algorithm, surpasses the performance of Lin \textit{et al.} \cite{lin2022deep}, a method specifically designed for vanishing point estimation. This outcome underscores the efficacy of the fine-level module of LNR in refining and assessing hypotheses. Furthermore, LNR achieves a processing speed of 10 Hz, outperforming most compared methods and delivering top-tier results, benefiting from GPU-accelerated parallel computing. Moreover, a reduced configuration, \emph{Ours-tiny}, achieves competitive real-time performance, demonstrating the efficiency of the multi-task loss training framework for robust multi-model fitting. While PARSAC \cite{kluger2024parsac} exhibits marginally faster speed, it requires prior knowledge of the number of hypotheses, a limitation in practical scenarios. Our proposed method, relying on accurate scoring, effectively fits multiple instances through HNMS without needing to predefine the number of hypotheses.

\begin{figure}[t]
  \centering
   \includegraphics[width=1.0\linewidth]{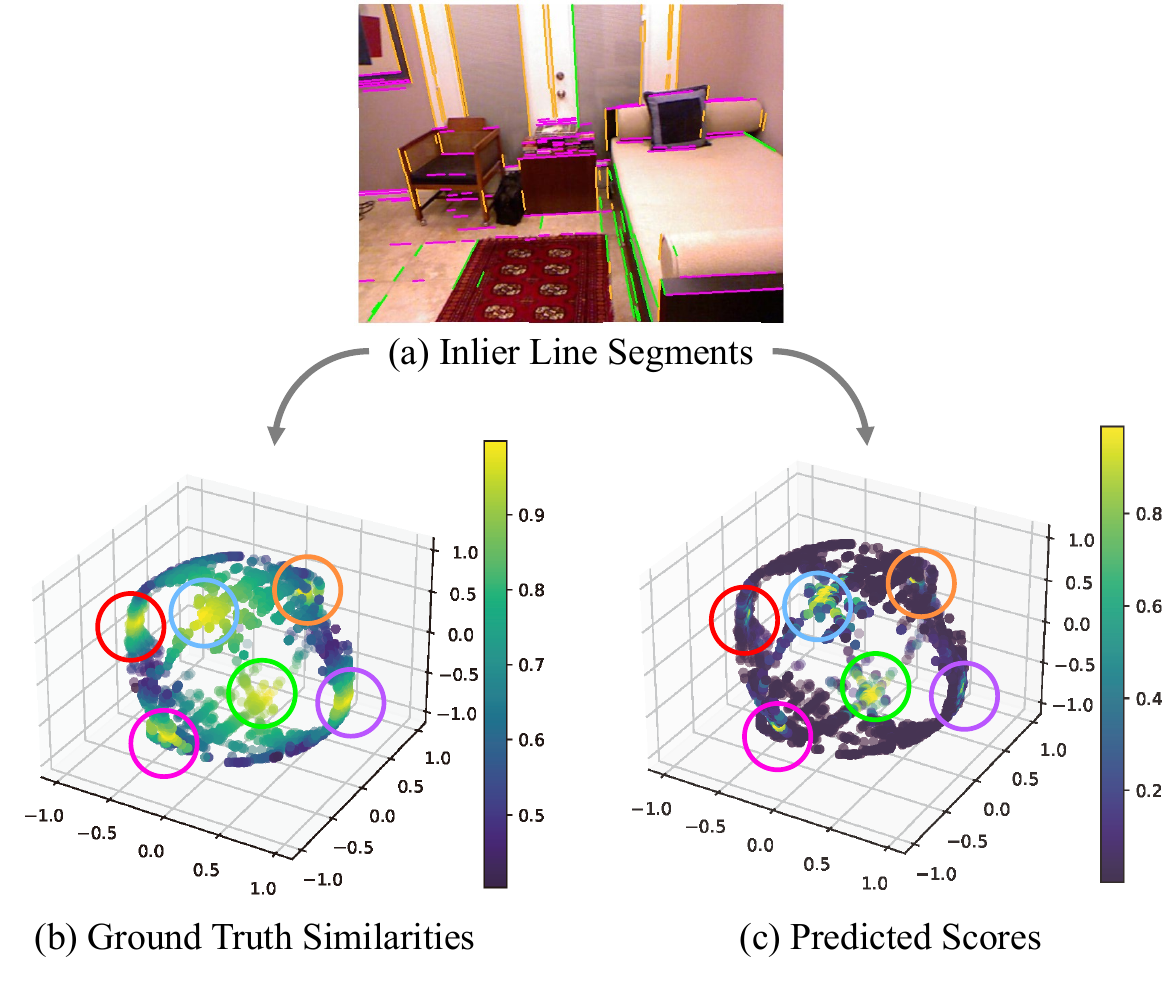}
    \vspace{-25pt}
   \caption{\textbf{Score visualization for the vanishing point estimation task.} (a) Line segments belonging to different vanishing points are indicated by different colors. All hypothetical vanishing points are projected on a Gaussian sphere (b) and (c). In (b), the linear similarities of the hypotheses to the ground truth are indicated by different colors. In (c), the scores for the hypotheses evaluated by LNR are also indicated by different colors. The same vanishing points in (b) and (c) are marked by the same color circles. LNR assigns high scores to correct hypotheses while greatly suppressing the scores of incorrect hypotheses, better than the linear similarities.} 
   \vspace{-10pt}
   \label{fig:vp_supp}
\end{figure}

TABLE \ref{tab:vptable} also reveals the sustained superior performance of LNR on the YUD+ and YUD datasets despite being trained only on NYU-VP. This highlights the exceptional generalization capability of LNR, validating the effectiveness of its region feature encoding. This suggests that our proposed method effectively classifies neighboring region points for refinement and assessment, enabling the selection of superior hypotheses from well-classified region points.

Qualitative evaluations in Fig. \ref{fig:vp} visualize vanishing point estimation performance across diverse scenes. Comparison with CONSAC\footnote{Implemented at \hyperlink{https://github.com/fkluger/consac}{https://github.com/fkluger/consac}.} \cite{kluger2020consac}, a state-of-the-art method, reveals that LNR more accurately determines line directions, resulting in fitting outcomes closer to the ground truth, especially for shorter lines. This directional accuracy contributes to the precise vanishing point estimation of LNR, consistently demonstrated across varied scenes. In contrast, CONSAC exhibits a tendency to misclassify and miss lines. The superior performance of LNR is attributed to the effective quality assessment scores and refined parameters of its generated hypotheses.

Score visualization for vanishing point estimation, presented in Fig. \ref{fig:vp_supp}, further underscore the capability of LNR to accurately determine line directions. The hypothetical vanishing points of the high scores assigned by the LNR are similar to the locations of the ground truth. Moreover, the LNR assigns low scores to bad hypotheses, providing an excellent suppression for these hypotheses to be selected. Such a property can make it easier for the correct hypotheses to be selected.

\subsection{Case Study 3: Two-view Plane Segmentation}

\begin{table}[t]\scriptsize 
\centering
\caption{\textbf{The quantitative result of two-view plane segmentation and two-view motion on AdelaideRMF dataset.} The average misclassification errors (avg. in \%, 5 runs) and their standard deviations (std.). The best indicators are bolded and sub-optimal indicators are underlined.}
\vspace{-7pt}
\setlength{\tabcolsep}{1.2mm}
\renewcommand\arraystretch{0.9}
\begin{tabular}{l||ccc|ccc}
\toprule
\multirow{2}{*}{Method}                                          & \multicolumn{3}{c|}{Two-view plane segmentation}                                               & \multicolumn{3}{c}{Two-view motion}                                                            \\ \cline{2-7} 
                                                                 & avg. ↓                        & std. ↓                        & Speed (Hz)↑                    & avg. ↓                        & std. ↓                        & Speed (Hz)↑                    \\ \hline
Prog-X \cite{barath2019progressive}             & 6.68                          & 5.84                          & 0.82                  & 10.72                         & 8.70                         & 0.92                           \\
PARSAC \cite{kluger2024parsac}                  & 7.24                          & 5.21                          & \textbf{12.31}                 & 8.73                          & 4.35                          & \textbf{60.13}                 \\
CONSAC \cite{kluger2020consac}                  & 5.23                          & 6.47                          & 0.11                           & -                             & -                             & -                              \\
Barath \textit{et al.} \cite{barath2023finding} & \underline{3.12}                 & 3.53                 & 1.59                           & \underline{5.33}                          & 4.42                          & 5.15                           \\
\textbf{Ours}                                                    & \textbf{2.91}                 & \textbf{3.14}                 & 4.79                           & \textbf{5.75}                 & \textbf{4.13}                 & 12.47                           \\
\textbf{Ours-tiny}                                               & \underline{3.12} & \underline{3.50} & \underline{10.21} & 6.27 & \underline{4.32} & \underline{30.26} \\ \bottomrule
\end{tabular}
\label{tab:Htable}
\vspace{-14pt}
\end{table}

Plane segmentation in two views involves detecting homographies between two images of the same scene to segment multiple 3D planes. For training, the HEB dataset \cite{barath2023large} is utilized, which provides 226,260 ground truth homographies. Given that a single image in HEB may contain multiple homographies, we merge HEB into a multi-homography dataset for training. Evaluation is conducted on the AdelaideRMF-H dataset \cite{wong2011dynamic}, employing the average misclassification error (ME) and its standard deviation as evaluation metrics, consistent with Prog-X \cite{barath2019progressive}. The input data, representing corresponding points across two views, is formulated as $\chi = \left\{ {\left[ {x_1^i,y_1^i,x_2^i,y_2^i} \right]\left| {\;i = 1,2,.... ,N} \right.} \right\}$, where $\chi \in \mathbb{R}^{N \times c}$ and $c=4$. LNR is evaluated against several competing methods.

The quantitative results for two-view plane segmentation, summarized in TABLE \ref{tab:Htable}. LNR achieves state-of-the-art performance, where the mean ME and standard deviation are slightly better than Barath \textit{et al.} \cite{barath2023finding}. Consistent with the vanishing point estimation task, the \emph{Ours-tiny} configuration of LNR provides notable speed enhancements. These results underscore the effectiveness of the encoded region features in both refining and scoring hypotheses, further illustrating its proficiency in handling image-based correspondences.

Qualitative evaluations, visualized in Fig. \ref{fig:H}, confirm the effectiveness of LNR in estimating 3D planes across diverse scenes. As one of the state-of-the-art methods, the method of Barath \textit{et al.}\footnote{It is implemented with \hyperlink{https://github.com/danini/clustering-in-consensus-space}{https://github.com/danini/clustering-in-consensus-space}. \label{Barath}} \cite{barath2023finding} is also visualized. Compared to Barath \textit{et al.} \cite{barath2023finding}, LNR is closer to the ground truth. For example, in contrast to Barath \textit{et al.} \cite{barath2023finding} which misses planes in columns 1 and 4 of Fig. \ref{fig:H}, LNR successfully segments all planes. This improvement is attributed to the ability of LNR to weight neighbors within a broader region, effectively mitigating noise interference. Furthermore, results in the last column of Fig. \ref{fig:H} indicate the applicability of LNR to single-model robust fitting scenarios. In these situations, Barath \textit{et al.} \cite{barath2023finding} incorrectly segments a single plane into two. In contrast, LNR accurately identifies the single plane, demonstrating the compatibility of our proposed method for the single model.

\subsection{Case Study 4: Two-view Motion}

\begin{figure*}[t]
  \centering
   \includegraphics[width=0.99\linewidth]{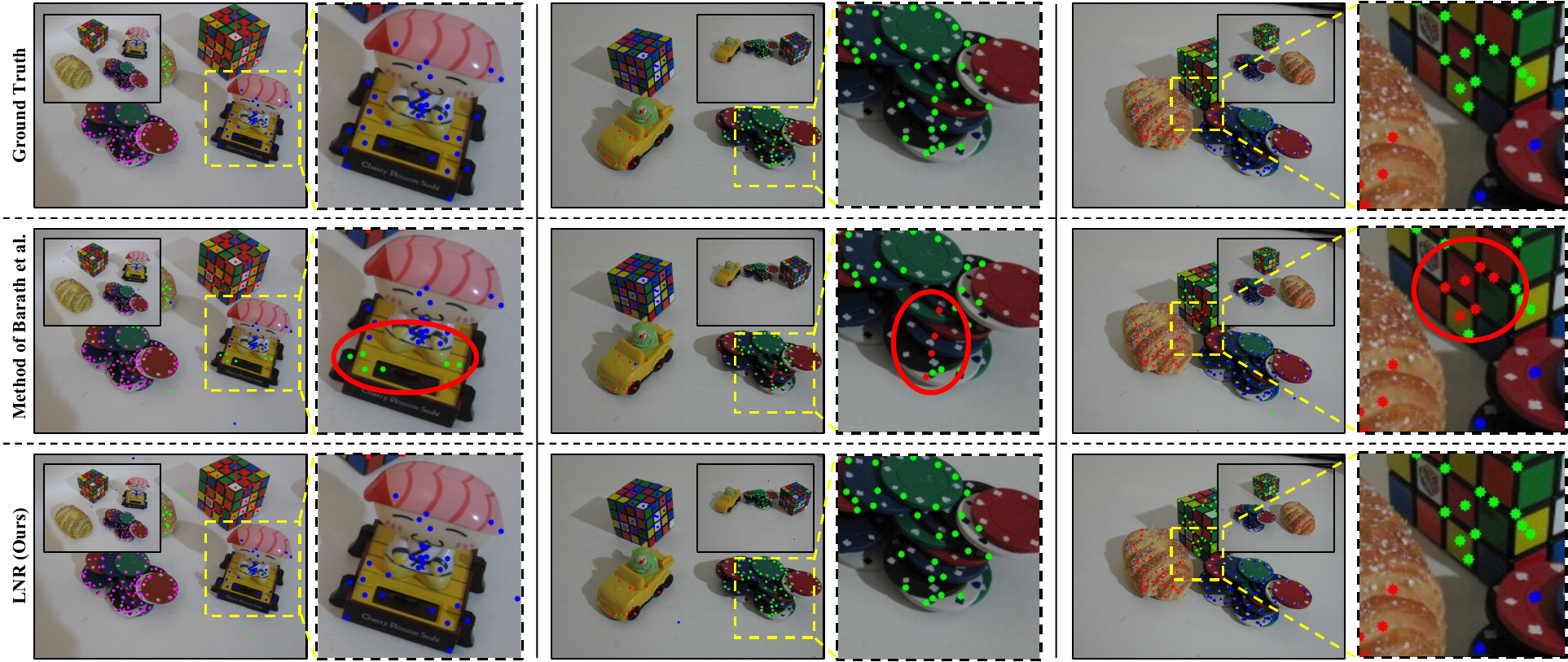}
    \vspace{-10pt}
   \caption{\textbf{The qualitative results of the method of Barath \textit{et al.} \cite{barath2023finding} and LNR on the two-view motion task in various scenes.} Different colors indicate which of the fundamental matrix the correspondences belong to. The small image and the large image are a pair of images. Their correspondences are used as input to the task. The major errors are indicated by red circles.} 
   \vspace{-14pt}
   \label{fig:F}
\end{figure*}

For two-view motion estimation, LNR is trained on the HOPE-F Dataset \cite{kluger2024parsac}, which comprises 4000 image pairs with keypoint features. Evaluation is performed using the AdelaideRMF-F dataset \cite{wong2011dynamic}, with the average ME and its standard deviation serving as the evaluation metrics. Input data representation $\chi \in \mathbb{R}^{N \times c}$ remains consistent with the two-view plane segmentation task, where $c=4$. our proposed method is compared with several other methods.

Quantitative results for two-view motion estimation, presented in TABLE \ref{tab:Htable}. LNR achieves state-of-the-art performance. Qualitative results, shown in Fig. \ref{fig:F}, further illustrate the superior ability of LNR to solve the ``point assignment to model" problem compared to the state-of-the-art method of Barath \textit{et al.}\textsuperscript{\ref{Barath}} \cite{barath2023finding}. This performance advantage is supported by LNR's efficient classification of data points based on geometric features.

\subsection{Ablation Study}
\label{ablation}
This section details an ablation study designed to evaluate the contribution of each module within the LNR framework. The experimental settings and evaluation metrics remain consistent with those previously established.

\begin{table}[t]\scriptsize 
\centering
\caption{\textbf{The ablation study of multi-line fitting.} The average AUC@$0.5^{\circ}$ ↑ on 1-10 line scenarios with five runs are reported. ``5" and ``6" indicate that the methods are trained only in ``5" and ``6" lines scenarios, tested directly on other scenarios. The best indicators are bolded and sub-optimal indicators are underlined.}
\vspace{-7pt}
\setlength{\tabcolsep}{0.75mm}
\renewcommand\arraystretch{1.0}
\begin{tabular}{l||cccccccccc}
\toprule
Method             & 1               & 2               & 3               & 4               & 5               & 6               & 7               & 8               & 9               & 10              \\ \hline \hline
Ours               & \textbf{96.73} & \textbf{93.41} & \underline{89.99} & 87.71 & \underline{84.89} & \textbf{83.32} & \textbf{81.47} & \textbf{79.91} & \textbf{78.59} & \textbf{77.51} \\
Ours-5             & \underline{92.33}          & \underline{91.82}          & \textbf{90.06} & \underline{87.75} & \underline{84.89} & \underline{82.58}          & 80.31          & 78.62          & 77.28          & 76.14          \\ 
Ours-6             & 91.25           & 91.47           & 89.83  & \textbf{87.90} & \textbf{85.47} & \textbf{83.32}          & \underline{81.41}          & \underline{79.67}          & \underline{78.37}          & \underline{77.20}         \\ \hline
CONSAC \cite{kluger2020consac} & \textbf{96.14}          & \textbf{91.29}          & \textbf{86.63}          & \underline{84.21}          & \underline{81.61}          & \textbf{79.83}          & \textbf{78.30}          & \textbf{76.34}          & \textbf{75.51}          & \textbf{74.84}          \\ 
CONSAC-5 \cite{kluger2020consac} & \underline{91.25}          & \underline{90.42}          & \underline{86.59}          & \textbf{84.23}          & \underline{81.61}          & \underline{79.12}          & 77.41         & 75.32          & 74.13          & 73.25          \\
CONSAC-6 \cite{kluger2020consac} & 90.32          & 90.01          & 86.21          & 84.15          & \textbf{81.64}          & \textbf{79.83}          & \underline{77.57}          & \underline{75.43}          & \underline{74.23}          & \underline{73.38}          \\ \bottomrule
\end{tabular}
\label{tab:line_ablation}
\vspace{-10pt}
\end{table}

\textbf{The generalization of LNR across different numbers of instances} is shown by \emph{Ours-5} and \emph{Ours-6} in TABLE \ref{tab:line_ablation}. Specifically, \emph{Ours-5} and \emph{Ours-6} denote models trained exclusively on scenes containing 5 and 6 lines, respectively. These models are then directly tested on scenes with different line counts. Performance metrics exceeding those of the standard \emph{Ours} model are highlighted in bold. Intriguingly, \emph{Ours-5} achieves performance levels closely comparable to \emph{Ours}. Even more notably, \emph{Ours-5} slightly surpasses the performance of LNR trained on scenes with 3 or 4 lines. This unexpected outcome suggests that training with a higher instance count, specifically 5 lines in this case, potentially introduces greater data variability or noise. This increased variability may foster a more robust model, enhancing the resilience of LNR to interference. Similar to \emph{Ours-5}, \emph{Ours-6} also exhibits a slight performance advantage over \emph{Ours} when evaluated on scenes with 4 and 5 lines. We performed the same setup on CONSAC \cite{kluger2020consac}. \emph{CONSAC-5} and \emph{CONSAC-6} exhibits the same findings, but its performance gains are smaller. Collectively, these results underscore the robustness and strong generalization capabilities inherent in our proposed method.

\begin{table}[t]\scriptsize 
\centering
\caption{\textbf{The ablation study results} of LNR for vanishing point estimation on the NYU-VP dataset. The AUC in \% is reported. The settings are consistent with TABLE \ref{tab:vptable}. The best indicators are bolded.}
\vspace{-7pt}
\setlength{\tabcolsep}{3.5mm}
\renewcommand\arraystretch{1.0}
\begin{tabular}{c|l||cc}
\toprule
                     &  Method                                            & @$5^{\circ}$↑                       & @$10^{\circ}$↑                \\ \hline \hline
\multirow{4}{*}{(a)} & Ours (w/o coarse-level)        & 49.26          & 63.12  \\
& Ours (pre-trained coarse-level)     & 55.14 & 69.42 \\ 
& \makecell[l]{Ours (full, multi-task loss \\trained coarse-level)} & \textbf{57.54} & \textbf{71.93} \\ \hline

\multirow{10}{*}{(b)} & \makecell[l]{Ours (w/o fine-level, coarse-level\\+inlier count)}        & 44.21          & 58.68 \\
& \makecell[l]{Ours (w/o fine-level, coarse-level\\+inlier count+inlier refinement)}   & 49.73          & 63.61 \\ 
& \makecell[l]{Ours (coarse-level\\+inlier count+fine-level refinement)}     & 52.41          & 67.21  \\
& \makecell[l]{Ours (coarse-level\\+fine-level count+inlier refinement)}     & 53.94          & 68.62 \\ 
& \makecell[l]{Ours (full, coarse-level\\+fine-level count+fine-level refinement)}        & \textbf{57.54}          & \textbf{71.93} \\ \hline
\multirow{3}{*}{(c)} & CONSAC-(Time:3.19s)  & 49.85          & 64.07 \\
& Ours (full, LNR)-\textbf{(Time:0.09s)}    & 57.54          & 71.93 \\ 
& CONSAC+LNR-(Time:3.32s)    & \textbf{57.63}   & \textbf{72.03}  \\ \hline
\multirow{3}{*}{(d)} & Ours (PointNet without Embedding)  & 57.31	&71.69 \\
& Ours (PointNet+Basic Embedding)    &57.42	  &71.81 \\ 
& Ours (full, PointNet+Fourier Embedding)    & \textbf{57.54}  & \textbf{71.93}  \\ \hline
\multirow{2}{*}{(e)} & Ours (LSQ refinement)         & 53.89          & 68.57 \\
                     & Ours (full, NN refinement) & \textbf{57.54}          & \textbf{71.93} \\ \bottomrule
\end{tabular}
\label{tab:ablation_study}
\vspace{-15pt}
\end{table}

\textbf{The effects of multi-task loss training} are examined in TABLE \ref{tab:ablation_study} (a). One approach considered is to directly predict hypotheses and refine them for numerous sampled minimum sets, bypassing the generation of coarse hypotheses altogether. However, removing the coarse-level module leads to a significant decline in performance. This degradation likely occurs because the network struggles to effectively learn from the substantially increased data volume associated with directly processing all minimum sets. Furthermore, the presence of excessive noisy data negatively affects the performance of the fine-level module in both scoring and refinement stages. While pre-training the coarse-level module is another possible strategy, it proves less effective than multi-task loss training. Multi-task training offers an advantage because the loss signals from the fine-level module are propagated back to optimize the coarse-level module, a mechanism observed in prior researches \cite{sun2021loftr, ma2018modeling}. This back-propagation enables the coarse-level module to prioritize the selection of coarse hypotheses that are more compatible with the fine-level module, rather than simply adhering to pre-defined labels. This synergistic interaction between the coarse-level and fine-level modules is facilitated by their shared input of encoded region features during training.

\textbf{The different coarse-to-fine approaches} are evaluated in TABLE \ref{tab:ablation_study} (b). This evaluation involves substituting the fine-level module with more conventional methodologies. The results indicate that employing a simple inlier count as a replacement yields inferior performance. This finding reinforces the conclusion from MQ-Net that ``consensus maximization does not favor the best model". Moreover, using inlier refinement as an alternative to the fine-level module also underperforms. Inlier refinement appears to be more susceptible to variations in data quality. These findings collectively validate the effectiveness and necessity of the dedicated fine-level module within the LNR framework.

\textbf{Combining the different sampling methods} are presented in TABLE \ref{tab:ablation_study} (c). LNR, in its standard configuration, employs an unbiased random sampling strategy to generate a large number of minimum sets, contrasting with probability-based sampling techniques like CONSAC. Substituting the random sampling with CONSAC sampling leads to improved performance. However, this performance gain comes at the cost of increased computational time, making the overall process considerably slower.

\textbf{The effects of Fourier feature mapping} is tested in TABLE \ref{tab:ablation_study} (d). Positional embedding, through Fourier feature mapping, converts geometric information into high-dimensional vectors. This transformation empowers the neural network to more effectively learn and represent subtle, high-frequency geometric details. Consequently, the network becomes better equipped to discern and learn the nuanced differences between individual data points.

\textbf{Different hypothesis refinement approaches} are tested in the TABLE \ref{tab:ablation_study} (e). The neural network refinement outperforms LSQ refinement. This is because the neural network effectively identifies and discards noisy points, enhancing the use of significant data points.

\textbf{The effects of Fourier feature mapping} is tested in TABLE \ref{tab:ablation_study} (d). The results demonstrate that neural network-based refinement surpasses traditional Least Squares (LSQ) refinement. This performance advantage can be attributed to the ability of neural network to effectively identify and disregard noisy data points, thereby prioritizing and leveraging the more informative and significant data points during the refinement process.

\vspace{-6pt}
\section{Discussion}
\label{disscussion}
\vspace{-4pt}
Traditional multi-model fitting methods, grounded in numerical computation, offer inherent interpretability and clearly defined application scopes. However, these approaches often struggle to fully exploit the geometric features of the data, potentially limiting their performance. Conversely, while deep learning techniques have demonstrated remarkable performance gains, end-to-end methods \cite{lin2022deep, chen2024vpdetr} that directly map input data points to multi-models can operate as ``black boxes", lacking transparency and potentially hindering their reliable deployment in practical applications.

In contrast, our proposed method seeks to bridge this gap by enhancing performance while maintaining interpretability. LNR leverages neural networks to effectively analyze the geometric features within the data. Crucially, for interpretability, LNR adopts the established two-step strategy: first, it gathers high-quality minimum sets of data points. Second, it applies minimum solvers to generate multi-models. This is achieved by training LNR to learn from data points within neighboring regions, rather than directly from hypotheses. This strategic choice naturally avoids the need to explicitly differentiate the sampling process and the model solvers, thus integrating traditional frameworks into the learning pipeline. Furthermore, Experiments indicate that learning-based hypothesis evaluation is significantly more efficient than most traditional rule-based evaluation techniques. Therefore, we hope to introduce a new idea for addressing multi-model fitting problems in learning fashions.

In addition, LNR can be easily transferred to various robust multi-model fitting tasks due to the following reasons:
\begin{itemize}
	\item Our proposed method accepts generic, unordered data points as input, encompassing various data modalities including image correspondences, 3D point clouds, and 2D pixel points, all uniformly represented as point sets.
 
	\item The network input focuses solely on the geometric features of these data points. While hypotheses and neighbor regions guide the utilization of these features, the loss function operates on the geometric features rather than the hypotheses. As a result, Our proposed method maintains compatibility with a broad spectrum of hypothesis solvers and sampling processes, whether differentiable or non-differentiable.

\end{itemize}

\vspace{-6pt}
\section{Conclusion}
\vspace{-4pt}
In this paper, we introduce LNR, a coarse-to-fine framework of Learning Neighbor Regions (LNR) to achieve robust multi-model fitting. For the inefficiency caused by sampling numerous minimum sets, LNR introduces a coarse-level module to select good minimum sets by analyzing geometric features, solving the coarse-hypotheses. For the model overlap and non-differentiable problem, LNR encodes neighbor region features for each coarse-hypothesis to refine and score these hypotheses individually. Since the region features consist of data points rather than hypotheses, the LNR avoids differentiating the sampling process and the model solvers. Furthermore, the multi-task loss function optimizes the coarse-to-fine process towards a consistent goal. In the experiments, the multi-line fitting task highlights the powerful generalization of LNR across different scenes. Notably, LNR excels in vanishing point estimation, two-view plane segmentation and two-view motion tasks, achieving state-of-the-art performance. LNR can effectively process correspondences and address the ``point-to-model assignment" problem. Significantly, LNR framework is not task-specific. It can be easily transferred to various robust multi-model fitting tasks.

\bibliographystyle{IEEEtran}
\bibliography{main}

\vfill

\end{document}